\pdfoutput=1
\documentclass[numbers,sort&compress]{IEEEtaes}
\usepackage{amsmath}
\usepackage{cite}
\usepackage{tabularx}  % 加载tabularx宏包以使用X列类型
\usepackage{color,array,amsthm}
\usepackage{graphicx}

\begin{document}

\bibliographystyle{unsrt}
\title{Energy Score-based Pseudo-Label Filtering and Adaptive Loss for Imbalanced Semi-supervised SAR target recognition}

\author{Xinzheng Zhang}
\member{Member, IEEE}
\affil{Chongqing University, Chongqing, China} 

\author{Yuqing Luo}
\affil{Chongqing University, Chongqing, China} 

\author{Guopeng Li}
\affil{Chongqing University, Chongqing, China} 

\pagestyle{empty}

%\receiveddate{Manuscript received XXXXX 00, 0000; revised XXXXX 00, 0000; accepted XXXXX 00, 0000.\\
%This work was supported in part by the National Natural Science Foundation of China under Grant 61301224 and in part by the National Natural Science Foundation of Chongqing under Grant cstc2021jcyjmsxmX0174. (Corresponding author: Xinzheng Zhang.)}
%% \accepteddate{XXXXX XX XXXX}
%% \publisheddate{XXXXX XX XXXX}

%\corresp{The name of the corresponding author appears after the financial information, e.g. {\itshape (Corresponding author: M. Smith)}. Here you may also indicate if authors contributed equally or if there are co-first authors.}

\authoraddress{Xinzheng Zhang is with the School of Microelectronics and Commu nication Engineering, Chongqing University, Chongqing 400044, China. and also with the Chongqing Key Laboratory of Space Information Net work and Intelligent Information Fusion, Chongqing 400044, China
(e-mail: \href{mailto:zhangxinzheng@cqu.edu.cn}{zhangxinzheng@cqu.edu.cn}).Yuqing Luo is with the School of Microelectronics and Communication Engineering, Chongqing University, Chongqing 400044, China
(e-mail: \href{mailto:vivian\_lyq@foxmail.com}{vivian\_lyq@foxmail.com}).Guopeng Li is also with the School of Microelectronics and Communication Engineering, Chongqing University, Chongqing 400044, China
(e-mail: \href{mailto:lgp\_junjue@163.com}{lgp\_junjue@163.com}).}
%
%%%\editor{Mentions of supplemental materials and animal/human rights statements can be included here.}
%%%\supplementary{Color versions of one or more of the figures in this article are available online at \href{http://ieeexplore.ieee.org}{http://ieeexplore.ieee.org}.}
%
%
%\markboth{Xinzheng Z ET AL.}{Energy Score-based Pseudo-Label Filtering and Adaptive Loss for Imbalanced Semi-supervised SAR target recognition}

\maketitle
\thispagestyle{empty}

\begin{abstract}Automatic target recognition (ATR) is an important use case for synthetic aperture radar (SAR) image interpretation. Recent years have seen significant advancements in SAR ATR technology based on semi-supervised learning. However, existing semi-supervised SAR ATR algorithms show low recognition accuracy in the case of class imbalance. This work offers a non-balanced semi-supervised SAR target recognition approach using dynamic energy scores and adaptive loss. First, an energy score-based method is developed to dynamically select unlabeled samples near to the training distribution as pseudo-labels during training, assuring pseudo-label reliability in long-tailed distribution circumstances. Secondly, loss functions suitable for class imbalances are proposed, including adaptive margin perception loss and adaptive hard triplet loss, the former offsets inter-class confusion of classifiers, alleviating the imbalance issue inherent in pseudo-label generation. The latter effectively tackles the model's preference for the majority class by focusing on complex difficult samples during training. Experimental results on extremely imbalanced SAR datasets demonstrate that the proposed method performs well under the dual constraints of scarce labels and data imbalance, effectively overcoming the model bias caused by data imbalance and achieving high-precision target recognition.
\end{abstract}


\section{INTRODUCTION}
S{\scshape ynthetic} Aperture Radar (SAR) is a high-resolution imaging radar that is frequently employed in homeland security since it can operate day and night\cite{1}. Automatic target recognition (ATR) technology can interpret complicated SAR target samples into valuable intelligence information\cite{2,3}, making it useful in a variety of military applications such as precise guidance and battlefield monitoring.

In recent years, the rapid growth of deep learning technology has brought about tremendous improvements in SAR ATR tasks\cite{4}. Zhou et al. introduced morphological operations to improve the quality of SAR target data. They designed a large-margin softmax batch-normalization CNN network structure to enhance the separability of samples after clutter removal\cite{5}. Zhang et al. fused the semantic features extracted by convolutional neural networks with traditional scattering center features, achieving a recognition accuracy of 99.59\% on the MSTAR dataset\cite{6}. Li et al. proposed a multiscale convolutional network and fully utilized the scattering centers of SAR targets to learn robust target features\cite{7}.

Deep learning algorithms offer higher prediction accuracy but larger model parameter sizes than conventional machine learning algorithms, and they require a significant amount of labeled data to support model training\cite{8,9}. However, manually labeling SAR images is costly, time-consuming, and error-prone. In recent years, various semi-supervised learning (SSL) systems have arisen, including Mean-Teacher\cite{10}, MixMatch\cite{11}, ReMixMatch\cite{12}, and FixMatch\cite{13}. These models have gained great success in the field of optical image identification, as well as offered strategies to address the label scarcity problem in SAR-ATR. Wang et al. introduced Mixup\cite{14} technique to combine labeled and unlabeled SAR data, which effectively utilizes unlabeled data and enhances SAR ATR performance\cite{15}. Yue et al. developed a semi-supervised SAR ATR framework that leverages labeled data for active learning and employs unlabeled data to impose constraints, yielding a notable improvement in recognition performance\cite{16}. Zhang et al. proposed a pseudo-label selection mechanism based on epoch and uncertainty sensitivity, which resulted in remarkable performance gains in scenarios with scarce labels\cite{17}. Zhang et al. designed two azimuth-aware discriminative representation losses that suppress intra-class variations among samples with large azimuth-angle differences, while simultaneously enlarging inter-class differences of samples with the same azimuth angle\cite{18}.

In general, the success of deep learning approaches is implicitly determined by the data scale's completeness. However, due to the unpredictable nature of wartime conditions and the high value of some prized military targets, certain unusual targets have characteristics such as excellent camouflage and mobility\cite{19}. This complicates the acquisition of their image data. Consequently, the lack of samples in some categories causes many SAR image datasets to exhibit an unbalanced data distribution\cite{20}, as shown in Figure 1, where some categories have abundant samples, referred to as the head categories, and data for other categories is scarce, termed as the tail categories. This objective contradiction creates a severe bottleneck for the SAR ATR technology. Under these circumstances, models are prone to severe overfitting difficulties, and their generalization capabilities deteriorate significantly\cite{21}.

\begin{figure}
\centerline{\includegraphics[width=18.5pc]{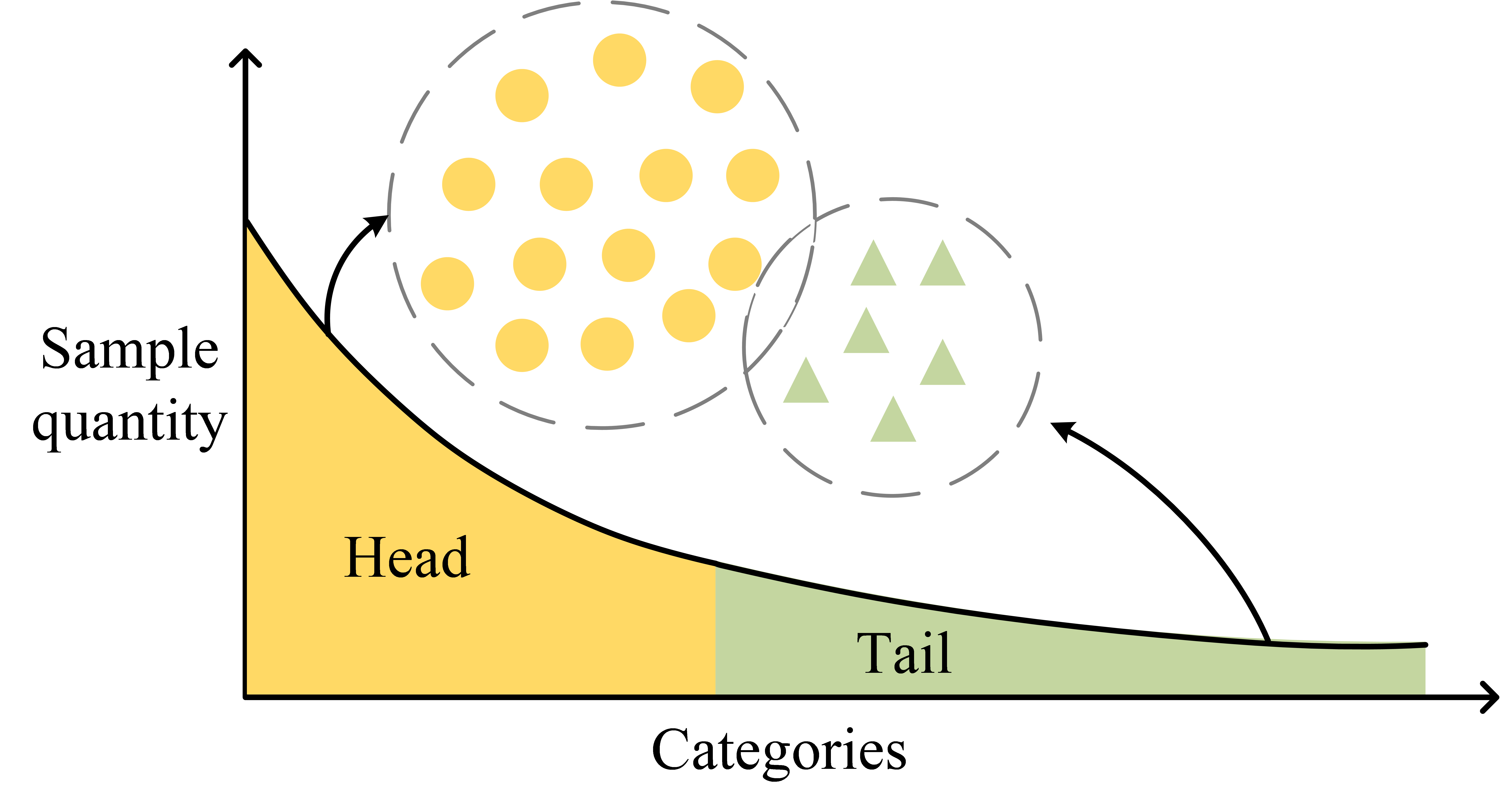}}
\caption{Long tail distribution diagram.}
\end{figure}

Existing strategies for dealing with imbalanced datasets in classification problems are broadly classified into two categories: data-level approaches and algorithm-level approaches. Data-level techniques preprocess the dataset without changing the network structure and are divided into five categories: oversampling\cite{22}, under-sampling\cite{23}, mixed sampling\cite{24}, feature selection\cite{25}, and deep generative models\cite{26}. Algorithm-level techniques do not preprocess the imbalanced dataset. Instead, they optimize the base classifier or adjust the network structure to improve the model's recognition rate. One common algorithm-level method assigns different costs to samples of different classes\cite{27,28,29}. For example, Li et al. proposed a cost-sensitive multi-decision tree method that alleviates the imbalance problem by imposing different penalties on different classes\cite{28}. Focal Loss reduces the weighting factor of accurately classified samples in the loss function, thereby enhancing the model's focus on the minority class samples\cite{29}. Ensemble learning has also been introduced to address data imbalance issues. Bhowmick et al. proposed a hybrid ensemble model based on the Boosting algorithm, which oversamples minority classes according to the recall rate and uses a voting algorithm for final classification\cite{30}. Transfer learning-based methods address class imbalance by modeling the majority and minority class samples separately and transferring the information learned from the majority class to the minority class\cite{31}. Metric learning aims to autonomously learn representative distance metric functions based on specific tasks to explore the similarities or differences between different class targets\cite{32}.

Current semi-supervised target recognition algorithms perform badly when data classes are imbalanced. This is because SSL implementations frequently use a high threshold for producing pseudo-labels to ensure correctness, which causes two issues. First, using high confidence exacerbates the imbalance problem by significantly decreasing the recall rate of pseudo-labels for data from minority classes\cite{33}. Second, samples far from the training data distribution might still have high prediction scores. This can lead to the incorrect classification of tail-class samples as high-confidence head-class samples\cite{34}, lowering the accuracy of recognition. Apart from the possible class imbalance found in SAR datasets, there is a high probability that the pseudo-labels produced will display a bias in favor of head classes\cite{35}, particularly in the first phases of training the model.

To address the aforementioned difficulties, this work investigates a non-balanced semi-supervised SAR ATR approach using dynamic energy scores and adaptive losses. First, we reject commonly used confidence-based pseudo-label generation methods and advocate the use of an Energy Score-based in-distribution Pseudo-label Selection (ESIDPS) mechanism that ensures the reliability of pseudo-labels in long-tailed data. Second, adaptive loss functions appropriate for imbalanced data are developed. On the one hand, to address the natural imbalance problem of pseudo-labels, we employ the Adaptive Margin Loss (AML) function rather than the generally used cross-entropy loss function in unsupervised losses. In addition, to address the model's preference for majority class samples, we propose an Adaptive Hard Triplet Loss (AHTL) function, which allows the model to focus on complicated hard samples and learn more discriminative feature representations.

Our main contributions are four-fold:
\begin{enumerate}
\def\labelenumi{\arabic{enumi})}
\item
  We introduce the concept of "within-distribution" and integrate it with the energy scores from the out-of-distribution detection task to enable quantitative selection. Throughout the training iteration process, the data distribution of pseudo-labels is constantly updated and expanded, boosting the reliability of pseudo-labels in scenarios of data imbalance.
\item
  We used an adaptive margin loss function to replace the traditional cross-entropy term in the unsupervised loss function. This was done to strengthen the model's awareness of the marginal differences between head and tail classes, thereby avoiding excessive bias towards majority classes and reducing the issue of biased pseudo-labels.
\item
  A self-adaptive hard triplet loss function was designed specifically for SSL. This loss function not only directs the model to focus more on complex and challenging samples during training, enhancing the network's feature learning capability and overcoming model bias but also adapts the weights of hard and easy samples dynamically. This adaptation prevents the model from being influenced by noisy samples and outliers, thereby improving the stability and robustness of the model.
\item
Extensive experiments were conducted on two sets of imbalanced SAR target datasets, MSTAR and FUSAR-ship. The experimental results demonstrate that the proposed method effectively enhances the recognition rate of SAR targets in class-imbalanced scenarios.
\end{enumerate}

The remaining sections of this paper are organized as follows. In Section II, the works related to this methodology are introduced. The proposed method is described in Section III. Section IV presents the experimental findings. Discussions of our method are included in Section V. In Section VI, this paper is finally concluded.

\section{Related work}

\subsection{SAR ATR Under Class imbalance}

In SAR target recognition tasks, the class imbalance problem poses a significant challenge. Researchers have put out several strategies to enhance recognition performance to overcome this problem. Data-level methods like oversampling\cite{22}, under-sampling\cite{23}, and the SMOTE technique\cite{36} balance the data by increasing the minority class samples or decreasing the majority class samples. Data augmentation techniques generate more minority class samples by rotating, translating, cropping, and other operations on SAR images. In recent years, generative adversarial networks, or GANs, have garnered a lot of attention as a new technique for synthesizing data. Cao et al. proposed a novel class-oriented GAN to augment imbalanced SAR datasets\cite{25}. This technique directs the balance by embedding a matrix into the model to reflect the generative model's requirement for various class samples.

Algorithm-level methods include cost-sensitive learning, which assigns higher weights to minority class samples during training, making the model more attentive to the minority class. Balanced loss functions such as Focal Loss, improves the model's ability to recognize minority classes by reducing the loss contribution from easily classified samples and increasing the loss contribution from hard-to-classify samples\cite{29}. Cui et al. proposed a class-balanced loss function based on the effective number of samples, effectively addressing the class imbalance problem in SAR image classification\cite{23}. Cao et al. addressed the issue of data imbalance from both data and algorithmic perspectives. They combined various oversampling methods to mitigate adverse correlations among target samples and employed a cost-sensitive model to alleviate the model's bias towards majority class samples\cite{37}.

Class imbalance has also been addressed through the use of transfer learning\cite{38,39,40}. Zhang et al. combined meta-learning and transfer learning and designed a cross-task and cross-domain SAR target recognition model. Which further enhanced recognition performance by incorporating domain confusion loss and a domain discriminator\cite{40}. Currently, little research on situations when labeled data is scarce is focused on class-imbalanced SAR ATR, most research is predicated on the assumption of sufficient SAR data labels.

\subsection{Pseudo-label Imbalance}

Pseudo-labeling is a widely used strategy in semi-supervised learning that generates pseudo-labels based on model predictions on unlabeled data, hence improving model performance by using more data. When generating pseudo-labels, state-of-the-art SSL techniques usually depend on confidence-based thresholds\cite{11,12,13}\cite{15}\cite{17}. Nonetheless, in situations involving long-tailed data distributions, these techniques frequently use higher confidence levels to enhance the precision of pseudo-labels\cite{33}. Sadly, this worsens the imbalance problem by drastically lowering the recall rate of pseudo-labels for minority class samples. Moreover, studies show that pseudo-labels may still display bias even when trained on balanced data\cite{41}.

The DASO model dynamically adjusts the contribution of each class's pseudo-label based on the class distribution, thus improving the recall rate of pseudo-labels for minority classes without sacrificing accuracy\cite{42}. Wang et al. first addressed the inherent imbalance issue in pseudo-labeling and proposed a debiased pseudo-label learning method\cite{41}, which eliminates the response bias of the classifier and adjusts the classification margins of each class. Yang et al. combined FixMatch and focal loss to alleviate the class imbalance issue in pseudo-labels\cite{43}, and accelerate model learning and convergence by adjusting loss weights based on predicted confidence.

\section{Method}

Figure 2 shows details of the proposed imbalanced semi-supervised SAR ATR approach. The method uses the EUAPS pseudo-label extraction process introduced by Zhang et al. as the first step to enlarge the original labeled dataset. Following that, experiments are run with the Wide ResNet 28-2 as the backbone network.

\begin{figure}
\centerline{\includegraphics[width=18.5pc]{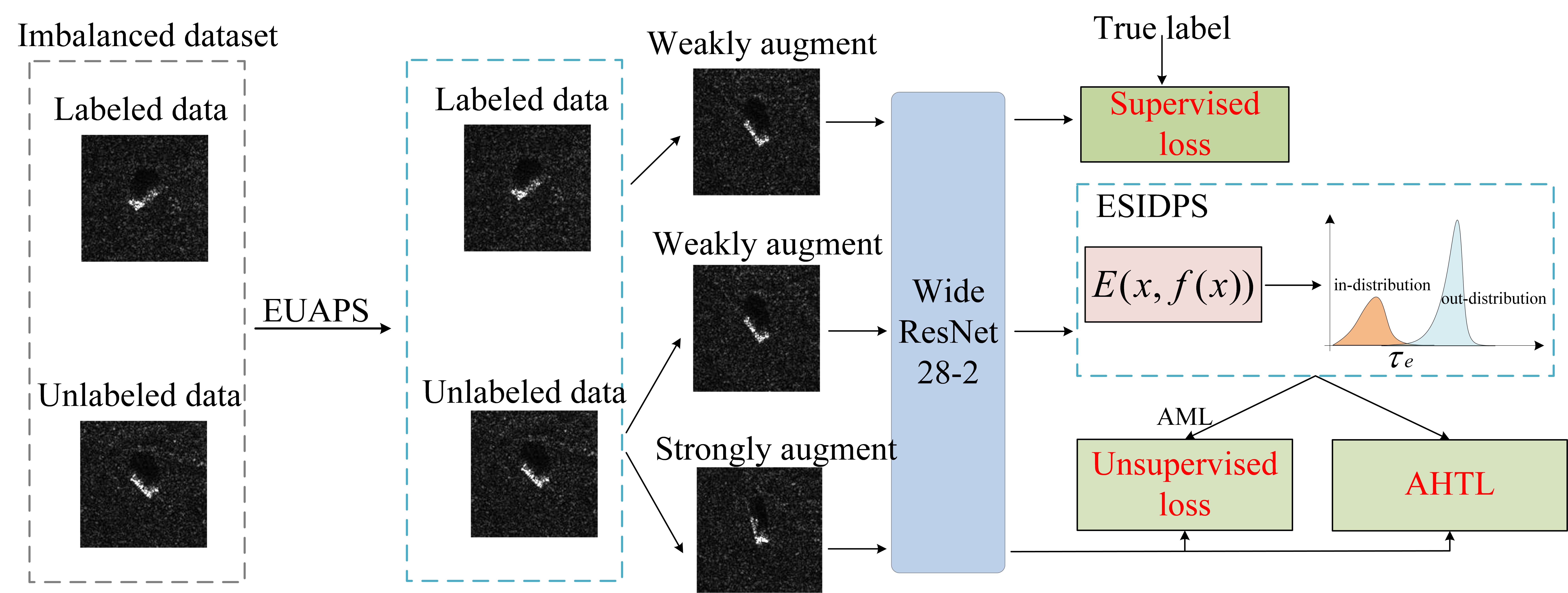}}
\caption{Overview of the proposed method.}
\end{figure}

\subsection{Energy Score-based in-distribution Pseudo-label Selection}

In SSL, confidence-based pseudo-label generating algorithms are frequently employed to ensure pseudo-label reliability by imposing a high threshold on low-confidence pseudo-labels. However, in the case of long-tailed distributions, such techniques drastically decrease the recall rate of minority class samples, boosting class imbalance problems. Furthermore, neural networks are overconfident, which means that samples misclassified by the classifier can nonetheless have high softmax prediction scores. This phenomenon may allow confidence-based pseudo-label selection approaches to misidentify some tail class samples as high-confidence head class samples, resulting in a drop in minority class recognition accuracy. To address these challenges, there is an urgent need to develop a pseudo-label selection technique suitable for long-tailed distribution scenarios.

Data is the foundation of machine learning, as models often investigate patterns and learn from them. However, in practical applications, the issue of out-of-distribution (OOD) generalization frequently emerges. Out-of-distribution data differs greatly from the model's previous observations, making it challenging for the model to produce accurate predictions for such data. In contrast, in-distribution refers to data that falls within the training distribution range, indicating that the model has already seen such data and is thus more likely to make accurate predictions. Machine learning tasks rely heavily on accurately detecting OOD data. One traditional method for OOD detection is based on softmax confidence; however, when the model overfits, such methods may assign a high confidence value to OOD samples, leading to incorrect detection. Another commonly used method is based on generative models; however, these methods face challenges such as difficulty in optimization and the need for prior density estimation of the data. To overcome these issues, Liu et al. suggested an energy score-based OOD detection method that produced satisfactory detection results. The energy score approach is not influenced by neural network overfitting, unlike softmax confidence methods, and it does not rely on density estimators, which avoids potential challenges in training generative models. As a result, this research proposes using energy scores instead of the softmax-based pseudo-label selection method.

In summary, the ESIDPS process in this work can be viewed as a continuously evolving in-distribution and out-of-distribution problem during training, as illustrated in Figure 3. Only a small number of initial labeled samples are considered in-distribution samples at the start of training because they have definite labels and supply the model with precise label information. Subsequently, at the end of each training iteration, energy scores for all unlabeled samples are recalculated, and samples that are close to the current training distribution are selected as pseudo-labeled samples and included in the in-distribution, gradually expanding the range of the in-distribution. This selection strategy assures that in-distribution samples have minimal uncertainty during each iteration, improving the model's reliability and stability. In addition, unlabeled samples that were previously designated out-of-distribution may be selected as in-distribution samples in subsequent iterations. This dynamic pseudo-label generation technique allows the model to better adapt to changing data distributions, which improves the model's generalization capacity.

The process of pseudo-label filtering is crucial as its quality directly impacts the subsequent model performance. This work aims to select samples that are close to the current training distribution, as they often have lower uncertainty, leading to a higher probability of correct predictions by the model. As shown in Figure 3, the yellow triangle sample on the left is far from the in-distribution area, indicating high uncertainty. However, approaches based on softmax confidence would classify it as a pseudo-labeled sample due to its high confidence level of 0.97. This demonstrates the limitations of softmax-based confidence approaches.

\begin{figure}
\centerline{\includegraphics[width=18.5pc]{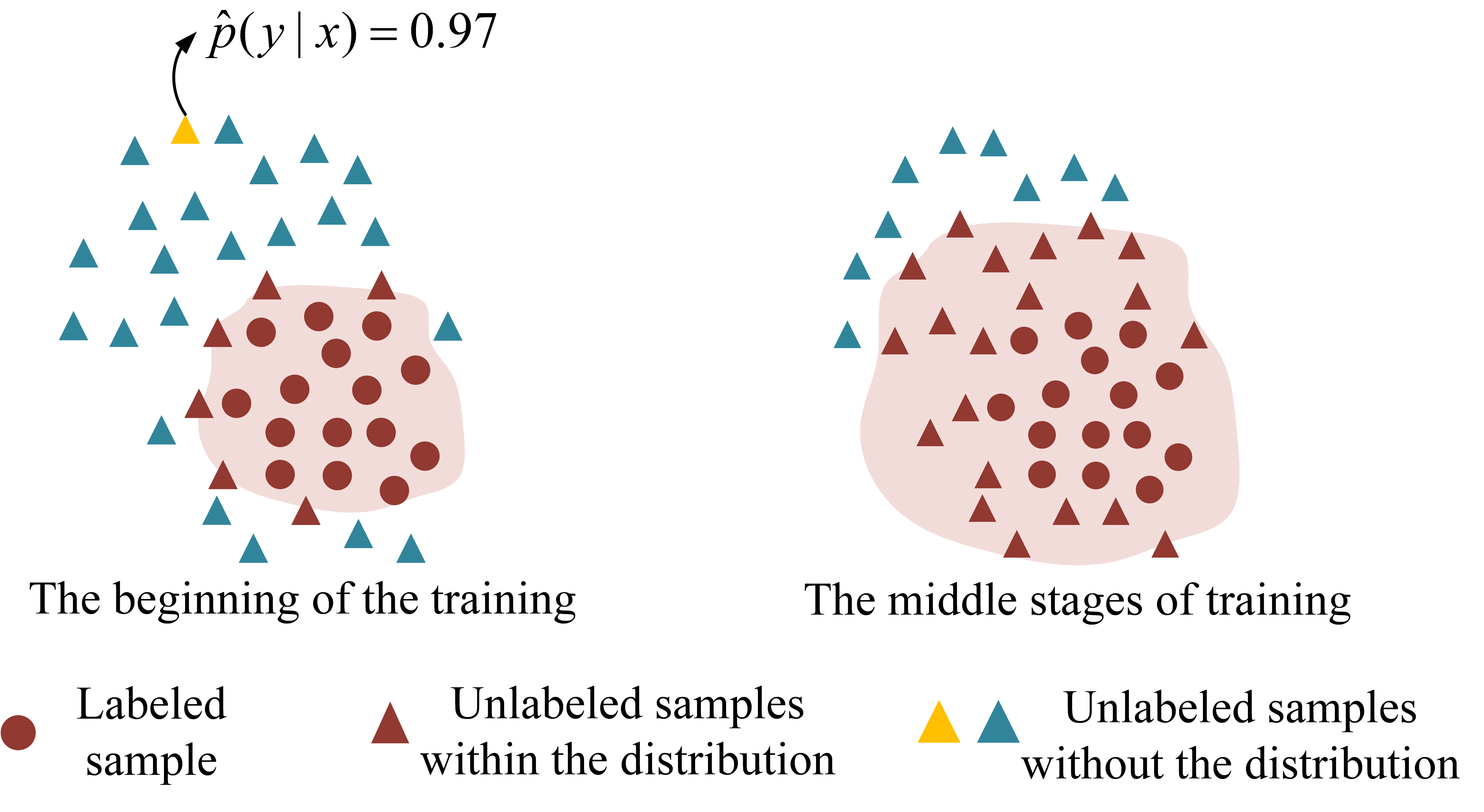}}
\caption{The evolution process of samples within and out of the distribution.}
\end{figure}

The energy score proposed by Liu et al. is a simple and efficient method. It is theoretically aligned with the probability density of the input data, where samples with lower energy scores are interpreted as having a higher likelihood of occurring, meaning they follow the current training distribution more closely. Conversely, higher energy scores indicate a lower likelihood of following the training distribution. The essence of computing energy scores is to establish an energy function $E(x){:}\mathbf{R}^D\to\mathbf{R}$, mapping each sample point in the input space to a scalar called energy. Through the Gibbs distribution, a set of energy values can be transformed into a probability density , expressed as:
\begin{equation}
p(y\mid x)=\frac{e^{-E(x,y)/T}}{\int_{y^{\prime}}e^{-E(x,y^{\prime})/T}}=\frac{e^{-E(x,y)/T}}{e^{-E(x)/T}}
\end{equation}

Where ${y}$ and $y^{\prime}$ correspond to the true label and predicted label of sample ${x}$, respectively. The denominator of the equation is called the partition function, which is used for normalization. ${T}$ is the temperature parameter used to smooth the probability distribution. Furthermore, the Helmholtz free energy ${E(x)}$ for a given input sample ${x}$ can be expressed as follows:
\begin{equation}
E(x)=-T\cdot\log(\int_{y^{\prime}}e^{-E(x,y^{\prime})/T})
\end{equation}

The energy score is closely related to the classifier $f(x){:}\mathbf{R}^D\to\mathbf{R}^K$. Let's define a discriminative classifier , which maps input sample to ${K}$ real numbers, known as logits. These logits can be used to derive the following probability distribution using the softmax function:
\begin{equation}
p(y\mid x)=\frac{e^{f_j(x)/T}}{\sum_i^Ke^{f_i(x)/T}}
\end{equation}

Here, ${f_j(x)}$ represents the element in ${f(x)}$ at index ${j}$, corresponding to the logit value for the ${j}$ class.

By combining equations (1) and (3), the energy score of input ${(x,y)}$ can be expressed as: ${E(x,y)=-f_y(x)}$ . Additionally, the energy function can be represented using the denominator of the softmax activation function. Thus, the following formula for computing the energy function is obtained:
\begin{equation}
E(x,f(x))=-T\cdot\log(\sum_{i=1}^Ke^{f_i(x)/T})
\end{equation}

The aforementioned energy calculation formula shows that the energy score is a simply implementable method. A lower energy value indicates that the sample is likely to be close to the current training distribution. Therefore, if the energy score of an unlabeled sample is lower than a predefined threshold $\tau_{e}$, the predicted class by the model can be considered as its pseudo-label and included in the training distribution. During each iteration of the model, the energy scores of each unlabeled sample are recalculated to dynamically update the pseudo-labels, enabling the model to adapt to the evolving data distribution. Figure 4 depicts the process of in-distribution pseudo-label selection based on energy scores.

\begin{figure}
\centerline{\includegraphics[width=18.5pc]{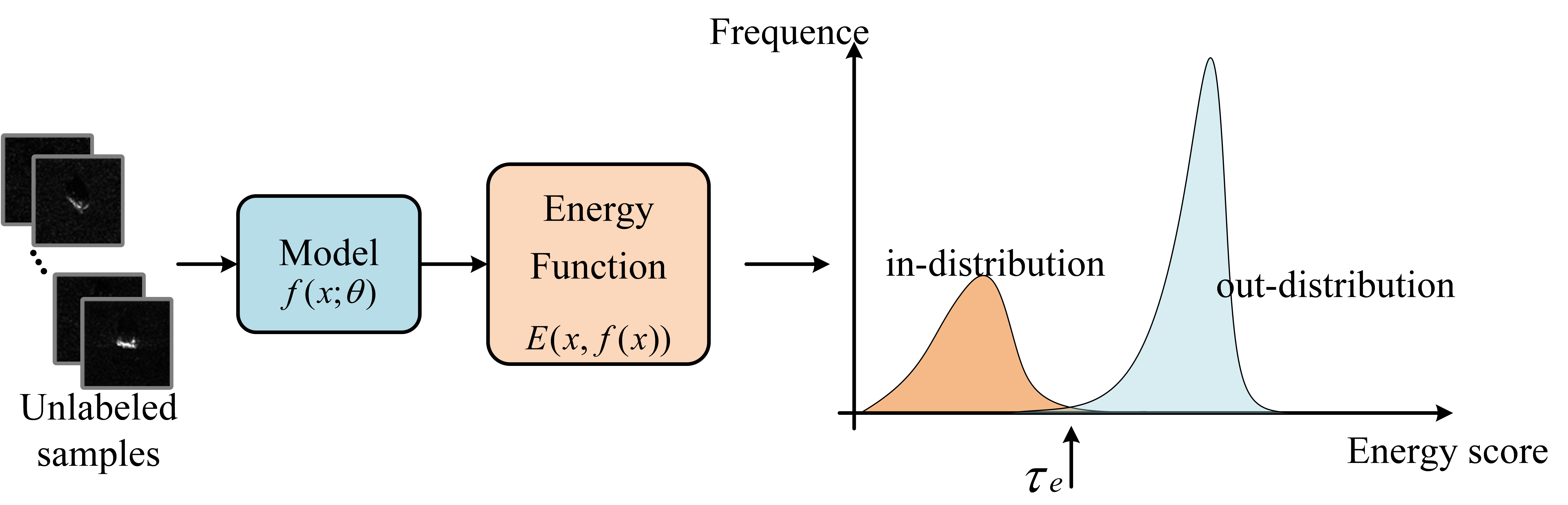}}
\caption{Energy score-based in-distribution pseudo-label selection method.}
\end{figure}

\subsection{Loss Term}

In SSL, objective recognition models typically comprise two loss function terms. One is the supervised loss ${L_s}$, computed on labeled data, and the other is the unsupervised loss computed on unlabeled data. The supervised loss ${L_s}$ is usually calculated to measure the discrepancy between the predicted labels of weakly augmented versions of labeled data and their true labels, and the formula can be written in the following form:
\begin{equation}
L_s=\frac1{N_X}\sum_{(x,y)\in X}H\left(y,p(\tilde{y}|w(x);\theta)\right)
\end{equation}

Where ${(x,y)}$ represents a sample and its corresponding label in the labeled dataset ${X}$, and ${w(x)}$ represents its weakly augmented version. ${N_X}$ represents the number of samples in the labeled dataset ${X}$, and $p(\tilde{y}|w(x);\theta)$ represents the predicted class probabilities by the ${\theta}$ parameterized model for sample ${w(x)}$.

The focus of semi-supervised learning research lies in designing the unsupervised loss function term to fully utilize the information in unlabeled data. Nowadays, the predominant approach is consistency regularization. It enhances the model's robustness by imposing consistency on the predictions of unlabeled samples subjected to different perturbations using a distance metric function. In recent years, SSL methods often introduce strong and weak augmentation to data to improve the model's generalization. Additionally, they employ thresholding and softmax confidence to select pseudo-labels. Assuming the weakly augmented version of unlabeled data ${u}$ is ${w(u)}$, and the model's predicted probabilities for ${w(u)}$ represented as $p(\tilde{y}|w(u);\theta)$, a method based on confidence selects samples with predicted probabilities higher than a confidence threshold $\mathcal{T}_{\mathcal{C}}$ as pseudo-labeled samples, with their one-hot labels denoted as $p(\tilde{y}|w(u);\theta)$. Then, a consistency loss function is constructed between the model's predictions on strongly augmented versions ${s(u)}$ and the pseudo-labels obtained on weakly augmented versions. Assuming the number of samples in the unlabeled dataset is ${N(U)}$, the form of the unsupervised loss function is as follows:
\begin{equation}
\begin{aligned}
&L_u=\frac1{N_U}\sum_{u\in U}1\Big[\max\big(p(\tilde{y} | w(u);\theta)\big)>\tau_c\Big]\\
&H\Big(\hat{p}\big(\tilde{y} | w(u);\theta\big),p(\tilde{y} | s(u);\theta)\Big)
\end{aligned}
\end{equation}

The shortcomings of pseudo-label generation methods based on softmax confidence were discussed, and the concept of energy scores was introduced to filter pseudo-labels close to the distribution. Therefore, the energy score of the weakly augmented version of unlabeled data ${w(u)}$ is calculated using formula (4), denoted as ${E(w(u),f(w(u)))}$. If the energy score is lower than a predefined threshold $\mathcal{T}_{\mathcal{e}}$, the model's predicted class for the sample is considered its pseudo-label. Thus, the unsupervised loss function term is rewritten from equation (4.6) to the following form:
\begin{equation}
\begin{aligned}
&L_u=\frac1{N_U}\sum_{u\in U}\mathbf{1}\Big[E\big(w(u),f(w(u))\big)<\tau_e\Big]\\
&H\Big(\Big(\hat{p}(\tilde{y} | w(u);\theta\Big),p_{model}(\tilde{y} | s(u);\theta)\Big)
\end{aligned}
\end{equation}

However, in practical applications, raw data that has not been preprocessed manually is mostly long-tailed distributed, making machine learning models prone to biases towards head classes. Therefore, it is necessary to consider imbalanced classification to prevent the classification model from being dominated by head classes. Additionally, research has shown that pseudo-labels generated in SSL tasks naturally suffer from imbalance issues, exacerbating the bias towards head classes. To address these issues, this paper improves the problem of model preference for majority classes from two aspects. Firstly, the cross-entropy term in the unsupervised loss function of semi-supervised learning is replaced by an adaptive margin loss to eliminate biases in the pseudo-label generation process. Secondly, an adaptive triplet loss function is designed to focus on hard samples, learning better feature representations. The following sections will provide detailed explanations of these two loss functions.

a.	Adaptive Margin Loss

The analysis above highlighted the inherent imbalance issue in the pseudo-label generation process. In such cases, the model is prone to generating biased pseudo-labels, and training the model with these incorrect labels further exacerbates the classifier's probability of misclassification. Wang et al. conducted in-depth research on the reasons for the imbalance in pseudo-labels in the FixMatch method, and through analysis of the correlation between different classes, they found that pseudo-labels for erroneous predictions often occur near the decision boundary. Therefore, biased pseudo-labels are largely attributed to inter-class confusion of the classifier. If unlabeled samples near the decision boundary can be pushed away from it, inter-class confusion can be effectively mitigated. Thus, it's beneficial to increase the distance between different classes in the feature space as much as possible, especially for classes with high similarity, to reduce the probability of generating incorrect pseudo-labels.

Based on these findings, this chapter introduces the Adaptive Margin Loss (AML) to dynamically adjust the margins of each class based on the imbalance degree of pseudo-labels. Specifically, it requires the model to maintain larger margins between highly biased and unbiased classes to ensure that the prediction scores of head classes do not overwhelmingly dominate over other classes. Such adjustments can effectively alleviate the imbalance issue caused by pseudo-labels while strengthening the model's discriminative ability across different classes. As shown in Figure 5, the classification boundary between Class 1 (yellow circle) and Class 2 (red triangle) in the left subplot is very close in the original feature space. AML aims to increase the boundary distance between these two classes to reduce inter-class confusion. Furthermore, throughout the training process, the data distribution constantly changes, so the bias between different classes should not be fixed but rather a dynamic process.

\begin{figure}
\centerline{\includegraphics[width=18.5pc]{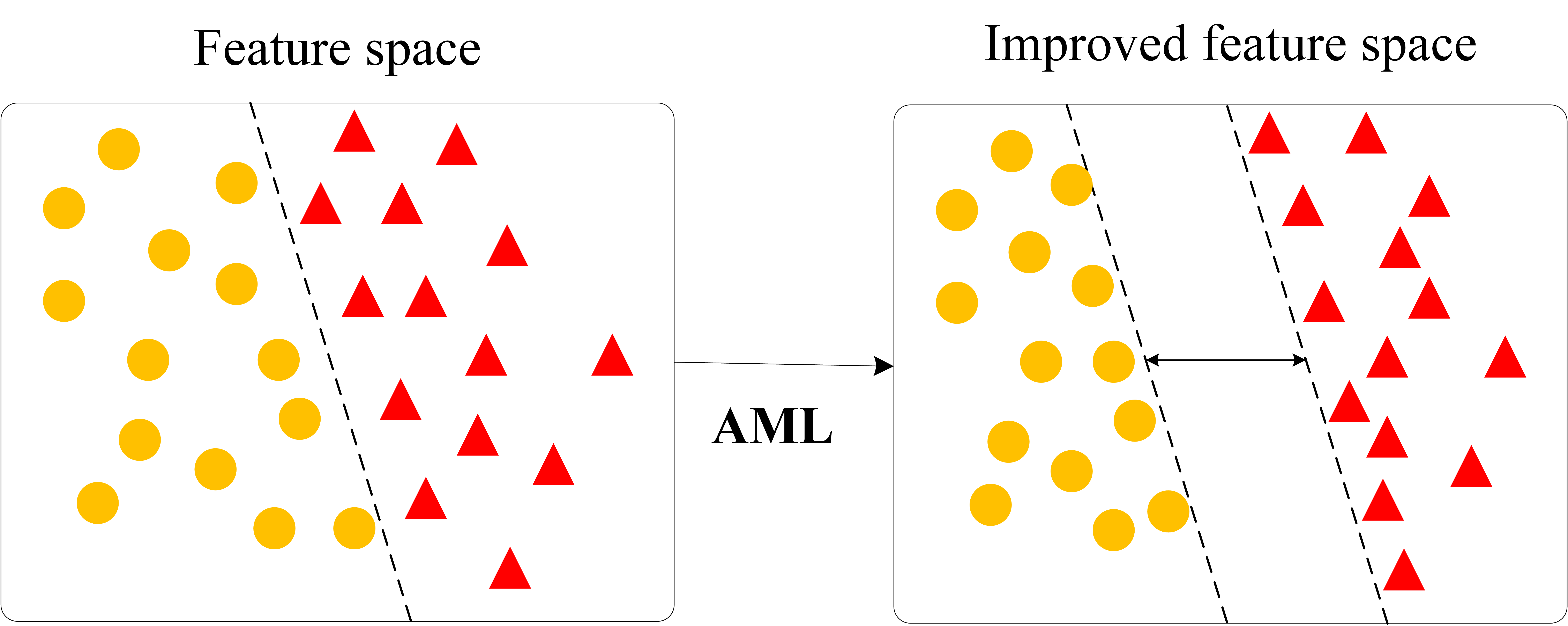}}
\caption{Adaptive margin loss diagram.}
\end{figure}

Before introducing the Adaptive Margin Loss function, let's briefly introduce the general softmax cross-entropy loss, which is widely used in fields like face recognition. It is composed of the softmax function and the cross-entropy loss, with the formula as follows:
\begin{equation}
L=-\sum_{i=1}^Ny_i\log S_i=-\frac1N\sum_{i=1}^N\log\frac{e^{f_i(x)}}{\sum_{k=1}^Ce^{f_k(x)}}\end{equation}

Here, ${S_i}$ represents the output probability value of the softmax function. ${y_i}$ denotes the true label of the sample, ${f_i(x)}$ represents the logits value for the classification of the sample, ${N}$ denotes the total number of samples, and ${C}$ represents the total number of classes.

The adaptive margin loss function used in this paper introduces some changes to the calculation of output probabilities using the softmax function compared to the regular softmax cross-entropy loss function. Specifically, AML adds an adaptively adjusted margin interval parameter ${m}$ at the position of the power. The core idea of AML is to dynamically adjust ${m}$ in the feature space, making the boundary distance between different classes larger to reduce inter-class confusion. For example, in a binary classification task, ${x_1}$ and ${x_2}$ represent the output vectors of two different classes after passing through the last fully connected layer, denoted as ${f(x_1),f(x_2)}$. To obtain a classifier with a larger inter-class margin, it is enforced that ${f(x_1),f(x_2)}$ satisfy    $f(x_1)-m_1>f(x_2)$ and $f(x_2)-m_2>f(x_1)$, where $m_1,m_2\geq0$ is used to adjust the inter-class margin in real-time. In this way, the classifier's ability to distinguish between different class features is strengthened, thereby improving overall classification performance. The specific form of ${L_{AML}}$  is as follows:
\begin{equation}
\begin{aligned}
&L_{AML}=-\frac1N\sum_{i=1}^N\log\frac{e^{(f_j(x_i)-m_j)}}{e^{(f_j(x_i)-m_j)}+\sum_{k\neq j}^Ce^{(f_k(x_i)-m_k)}} \\
&where\quad m_j=\lambda\log(\frac1{\hat{p}_j}),j\in\{1,...,C\}
\end{aligned}
\end{equation}

In the equation, ${f_j(x_i)}$ represents the logits value for sample ${x_i}$ classified as class ${j}$, $k\in[1,C]$. ${N}$ and ${C}$ respectively denote the total number of samples and the total number of classes. ${m_j}$ is the adaptively adjusted margin parameter calculated based on the predicted probabilities of different classes. $\hat{p}_j$ is the average model prediction value updated by exponential moving average at each iteration. When a sample belongs to the majority class, its contribution to the loss is reduced by the adaptive margin, thus preventing the majority class from suppressing the minority class. Finally, the $L_{AML}$ loss is used instead of the unsupervised loss, replacing the cross-entropy term in equation (7), to mitigate the imbalance issue in the pseudo-label generation process.

b. Adaptive Hard Triplet Loss

In scenarios of data imbalance, machine learning models tend to be biased towards the majority class. One approach to overcome this bias is Deep Metric Learning (DML). DML is a category of algorithmic methods that utilize neural networks to map samples into an embedding space, continuously reducing intra-class distances and increasing inter-class distances in this space to address the model's bias issue. Among them, Triplet Loss is the most representative method in DML, initially proposed by Google researchers and achieving significant success in the field of face recognition. It operates on triplets of samples, as illustrated in Figure 6, where each triplet comprises three samples: an anchor sample, a positive sample, and a negative sample. The anchor and positive samples belong to the same class, while the negative sample is from a different class. Triplet Loss aims to learn a feature space where the distance between the anchor and positive samples is minimized, and the distance between the anchor and negative samples is maximized. By optimizing this feature space, it ensures that samples from the same class are close to each other while samples from different classes are far apart, thus enabling a robust classifier. The general form of Triplet Loss is represented as follows:
\begin{equation}
\begin{aligned}
L_{triplet} = \sum_{i=1}^N \max \biggl[ & D(f(x_i^a), f(x_i^p)) - \\
& D(f(x_i^a), f(x_i^n)) + m, 0 \biggr]
\end{aligned}
\end{equation}

Here, ${x_i^a,x_i^p,x_i^n}$ represent the anchor, positive, and negative samples, respectively.   is the network mapping function, $f(\cdot)$ is the distance metric function, commonly used are Euclidean distance and cosine distance. $D(\cdot)$ is a hyperparameter denoting the margin, which is the minimum distance between $x_i^a$ and $x_i^p$ compared to $x_i^a$ and $x_i^n$. The margin $m$ is a constant greater than 0, and setting an appropriate value is crucial; too small values cannot enhance the model's discrimination ability among different classes, while too large values increase training difficulty and may lead to non-convergence of the model.

\begin{figure}
\centerline{\includegraphics[width=18.5pc]{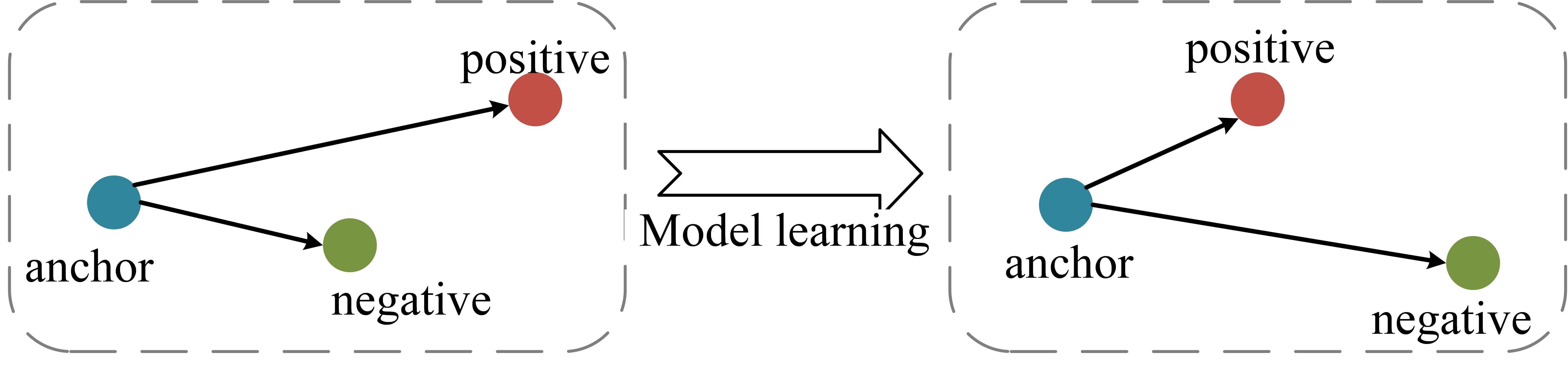}}
\caption{Triplet diagram.}
\end{figure}

The triplet loss function can learn to distinguish between two inputs with small differences, capturing finer feature details, which makes it perform well in image classification tasks. However, traditional triplet loss randomly samples triplets during construction, and most of the sampled triplets consist of relatively easy-to-distinguish samples. While this accelerates the convergence of the model, it also leads the model to ignore learning from more complex samples, hindering the model from learning better feature representations.

To address this issue, this chapter introduces the concept of Hard Triplet Loss to enhance the model's learning ability for latent features. Specifically, in constructing triplets, Hard Triplet Loss selects the sample with the lowest similarity to the anchor sample from the same class as the hardest positive sample. Similarly, it selects the sample with the highest similarity to the anchor sample from a different class as the hardest negative sample. By introducing the mechanism of hard samples, Hard Triplet Loss enables the model to focus more on learning challenging samples, thereby improving the model's applicability to complex tasks and its generalization in real-world scenarios. Additionally, this study adaptively assigns different weights to samples, allowing the model to focus on learning hard samples. As this paper's model is based on semi-supervised learning and lacks pre-existing label information, an adaptive Hard Triplet Loss function is designed specifically for this process. Here's an elucidation of this process:

Firstly, pseudo-labels filtered by energy scores and adaptive margin loss are highly accurate. Hence, hard triplets can be constructed from these pseudo-labels. Specifically, the pseudo-labels filtered by energy scores are used as anchor samples, represented by their corresponding feature vectors $\mathcal{W}_i^a$. For each anchor sample, the corresponding hard positive and hard negative samples are found from the pseudo-labels based on their similarity, where Euclidean distance is employed as the similarity metric. Then, the feature vectors of the corresponding strong augmentation versions of unlabeled samples are retrieved as representations $\mathcal{s}_i^p$ and $\mathcal{s}_i^n$. This constructs a triplet unit $<w_i^a,s_i^p,s_i^n>$. By using weak augmentation of unlabeled samples as anchor samples and strong augmentation as positive and negative samples, AHTL applies additional constraints to samples under different perturbations, aiding in improving the model's generalization and robustness.

Moreover, considering that traditional triplet and hard triplet loss functions typically assign uniform weights to all samples, making them vulnerable to noise and outliers, this study designs an adaptive triplet loss as shown in Equation (11), dynamically adjusting the weights of samples based on the feature distance between anchor and positive/negative samples.
\begin{equation}
\begin{aligned}
L_{AHTL} = \sum_{i=1}^N \max \Big[ &w_p \cdot \left\| w_i^\alpha - s_i^p \right\|_2^2 - \\
& w_n \cdot \left\| w_i^\alpha - s_i^n \right\|_2^2 + m, 0 \Big]
\end{aligned}
\end{equation}

The formula for calculating the weight parameter is as follows:
\begin{equation}
\begin{aligned}
&w_{p}=\frac{e^{d(w_i^a,s_i^p)}}{\sum_ie^{d(w_i^a,s_i^p)}}\\
&w_{n}=\frac{e^{-d(w_i^a,s_i^n)}}{\sum_ie^{-d(w_i^a,s_i^n)}}
\end{aligned}
\end{equation}

The formula shows that when the feature distance between the anchor $x_i^a$ and the positive sample $s_i^p$ is larger, the assigned weight is greater, indicating that the model will focus more on learning such samples. For the negative sample $s_i^n$, the weight is calculated using the negative softmax. Therefore, when the feature distance between the negative sample and the anchor is smaller, the assigned weight becomes larger. Through this adaptive weight adjustment mechanism, the model can better adapt to various complex data distributions and noise conditions, thereby improving training stability, accelerating model convergence, and enhancing overall performance.

Finally, the total loss function of the model consists of three parts, as shown in the formula below:
\begin{equation}
L=L_s+\lambda_uL_u+\lambda_{AHTL}L_{AHTL}
\end{equation}

Where $\lambda_{u}$ and $\lambda_{AHTL}$ represent the weights of the loss terms, used to balance the supervised and unsupervised losses.

\section{Numerical Experiments}

\subsection{Introduction to datasets}

In the practical application of SAR ATR, class imbalance is a challenge faced by many practitioners. The imbalance ratio (IR) typically represents the ratio of the number of samples in the majority class to the number of samples in the minority class in a dataset. The effect of different IR values on target recognition models varies. Generally, an IR value exceeding 9 indicates a highly imbalanced dataset. In this chapter, we select the representative MSTAR [9] and FUSAR-ship datasets [87] to validate the performance of the proposed method on imbalanced data.

(1)	MSTAR

The MSTAR dataset is obtained through a high-resolution, coherent X-band radar sensor with an imaging resolution of 0.3 m × 0.3 m, using the HH polarization mode. It contains images of various categories of static vehicle targets. The commonly used MSTAR dataset consists of ten classes, including the following target categories: 2S1, BMP2, BRDM2, BTR60, BTR70, D7, T62, T72, ZIL131, and ZSU234. The images of each target class are distributed at intervals of 1-5° in azimuth angle. Additionally, all target images are sampled at 15° and 17° elevation angles. Typically, images at a 17° elevation angle are used as the training dataset, while those at a 15° elevation angle are used as the testing dataset.

To obtain imbalanced datasets, we set the IR values to 10, 20, and 30, respectively. Following the work of previous researchers [97], we construct imbalanced datasets that follow a long-tailed distribution. Assuming the total number of samples in the majority class is N, and K is the total number of classes in the dataset, the formula for calculating the number of samples for class k when the imbalance ratio is IR is shown in equation (14). Due to the inconsistent sizes of the original data images and the concentration of vehicle targets mainly in the central range of the images, all images are cropped to the center to obtain slices of size 128 × 128. Figure 7 illustrates the SAR images of all classes in the MSTAR dataset along with their corresponding optical images. 
Table 1 provides detailed information on the number of samples for each class in the imbalanced MSTAR training set under different IR values.
\begin{equation}
N_k=N\cdot IR^{-(k-1)/(K-1)}
\end{equation}

\begin{figure}[h]
 \centering % 使所有的 minipage 居中
 \begin{minipage}{0.16\linewidth}
 	\vspace{1pt}
 	\includegraphics[width=\textwidth]{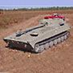}
 	\vspace{1pt}
 	\includegraphics[width=\textwidth]{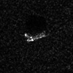}
 	\vspace{1pt}
 	\centerline{(a)}
 \end{minipage}
 \begin{minipage}{0.16\linewidth}
 	\vspace{1pt}
 	\includegraphics[width=\textwidth]{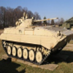}
 	\vspace{1pt}
 	\includegraphics[width=\textwidth]{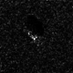}
 	\vspace{1pt}
 	\centerline{(b)}
 \end{minipage}
 \begin{minipage}{0.16\linewidth}
 	\vspace{1pt}
 	\includegraphics[width=\textwidth]{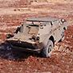}
 	\vspace{1pt}
 	\includegraphics[width=\textwidth]{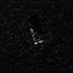}
 	\vspace{1pt}
 	\centerline{(c)}
 \end{minipage}
 \begin{minipage}{0.16\linewidth}
 	\vspace{1pt}
 	\includegraphics[width=\textwidth]{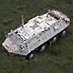}
 	\vspace{1pt}
 	\includegraphics[width=\textwidth]{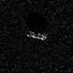}
 	\vspace{1pt}
 	\centerline{(d)}
 \end{minipage}
 \begin{minipage}{0.16\linewidth}
 	\vspace{1pt}
 	\includegraphics[width=\textwidth]{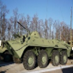}
 	\vspace{1pt}
 	\includegraphics[width=\textwidth]{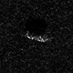}
 	\vspace{1pt}
 	\centerline{(e)}
 \end{minipage}
 \begin{minipage}{0.16\linewidth}
 	\vspace{1pt}
 	\includegraphics[width=\textwidth]{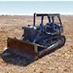}
 	\vspace{1pt}
 	\includegraphics[width=\textwidth]{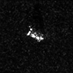}
 	\vspace{1pt}
 	\centerline{(f)}
 \end{minipage}
 \begin{minipage}{0.16\linewidth}
 	\vspace{1pt}
 	\includegraphics[width=\textwidth]{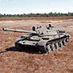}
 	\vspace{1pt}
 	\includegraphics[width=\textwidth]{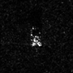}
 	\vspace{1pt}
 	\centerline{(g)}
 \end{minipage}
 \begin{minipage}{0.16\linewidth}
 	\vspace{1pt}
 	\includegraphics[width=\textwidth]{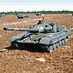}
 	\vspace{1pt}
 	\includegraphics[width=\textwidth]{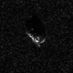}
 	\vspace{1pt}
 	\centerline{(h)}
 \end{minipage}
 \begin{minipage}{0.16\linewidth}
 	\vspace{1pt}
 	\includegraphics[width=\textwidth]{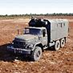}
 	\vspace{1pt}
 	\includegraphics[width=\textwidth]{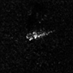}
 	\vspace{1pt}
 	\centerline{(i)}
 \end{minipage}
 \begin{minipage}{0.16\linewidth}
 	\vspace{1pt}
 	\includegraphics[width=\textwidth]{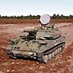}
 	\vspace{1pt}
 	\includegraphics[width=\textwidth]{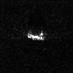}
 	\vspace{1pt}
 	\centerline{(j)}
 \end{minipage}
\caption{Optical images (top) and the corresponding SAR images (bottom) of the targets in MSTAR dataset.
(a)2S1; (b)BMP2; (c)BRDM2; (d)BTR60; (e)BTR70; (f)D7; (g)T62; (h)T72; (i)ZIL131; (j)ZSU234
.}
\end{figure}

\begin{table}
\caption{The samples class and number of samples for each category in the MSTAR training set under different IR values}
\label{table1}
\tablefont
\begin{tabular*}{20pc}{@{}c@{}}
\centerline{\includegraphics[width=18.5pc]{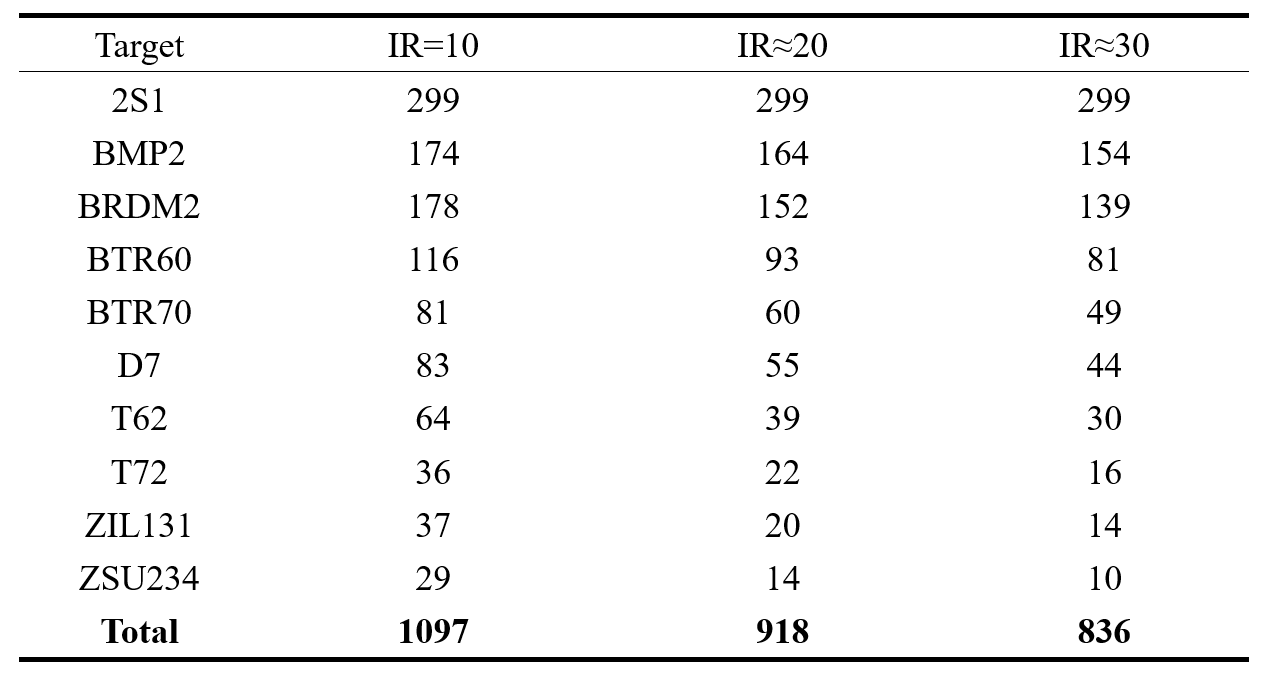}}\\
\end{tabular*}
\label{tab1}
\end{table}

(2)	FUSAR-ship

In 2020, Hou et al. released the FUSAR-ship dataset collected by the Gaofen-3 satellite. Gaofen-3 is China's first civilian fully polarimetric C-band synthetic aperture radar (SAR) satellite, with a resolution of up to 1 meter. FUSAR-ship is an open SAR-AIS co-registered dataset containing matching information on time, space, and coordinate transformation. The azimuth resolution of these target images is 1.124 m, and the range resolution ranges from 1.700 m to 1.754 m. The FUSAR-ship dataset exhibits an issue of data imbalance. In the experiments, five classes of ship targets were selected for training and testing: bulk carriers, oil tankers, fishing boats, container ships, and general cargo ships. The ratio between the majority and minority classes in this dataset is 10, making the IR equal to 10. Similarly, images were cropped from the center of the original images to a size of 96 × 96 as input for the model. Figure 8 displays the SAR images of these five types of ship targets, and Table 3 lists detailed information about the training and testing sets.

\begin{figure}
\centerline{\includegraphics[width=18.5pc]{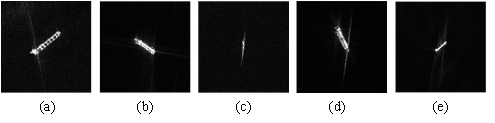}}
\caption{Triplet diagram.}
\end{figure}

\begin{table}
\caption{The samples class and number of training and testing set used in the FUSAR-ship dataset}
\label{table2}
\tablefont
\begin{tabular*}{20pc}{@{}c@{}}
\centerline{\includegraphics[width=18.5pc]{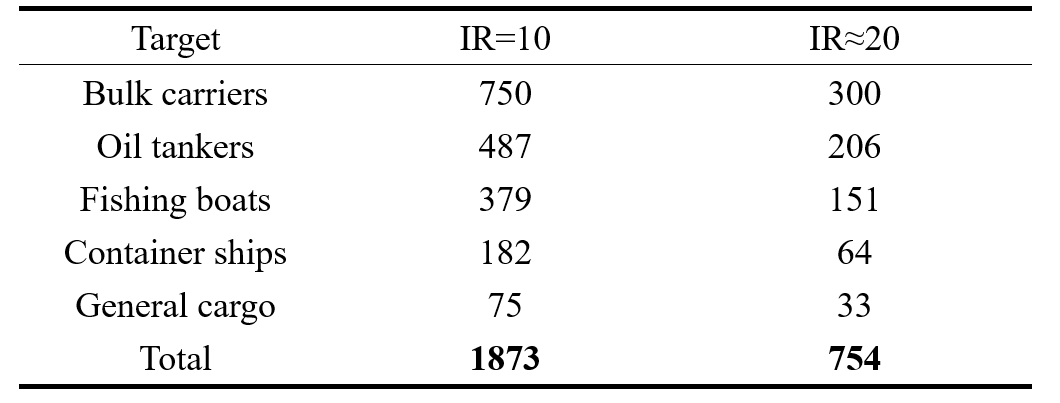}}\\
\end{tabular*}
\label{tab2}
\end{table}

\subsection{Implement details }

This chapter proposes a model that uses Wide ResNet 28-2 as the backbone network, with network parameters referencing FixMatch. The learning rate of the model is set to 0.03. SGD is chosen as the optimizer, with a momentum of 0.9. Within each training batch, there are 16 labeled samples, and the ratio of unlabeled samples to labeled samples is 7:1. By using more unlabeled data within each training batch, the training performance of the model can be effectively improved. For the MSTAR and FUSAR-ship datasets, the threshold for the energy score $\tau_{e}$ in equation (7) is set to -9.5 and -9, respectively. The temperature parameter T in equation (4) is set to 1 and 0.5 for the two datasets, respectively. Similarly, for the above two datasets, the hyperparameter   in equation (11) is set to 0.3. The weights $\lambda_{u}$ and $\lambda_{AHTL}$ for the unsupervised loss in equation (13) are set to 1.0 and 1.5, respectively. Test experiments are conducted using the exponential moving average of model parameters.
All experiments are implemented on a personal computer with an Intel Core i7-7700K CPU and 16GB of memory. The computer is equipped with a GeForce GTX 1080Ti GPU with 11GB of memory and PyTorch 1.9.0.

\subsection{Comparative experiments}

1) MSTAR dataset:

\begin{table}[h]
\caption{Comparison of recognition accuracy (\%) between the proposed method and other algorithms on the MSTAR dataset, best results in bold.}
\label{table3}
\tablefont
\begin{tabular*}{20pc}{@{}c@{}}
\centerline{\includegraphics[width=18.5pc]{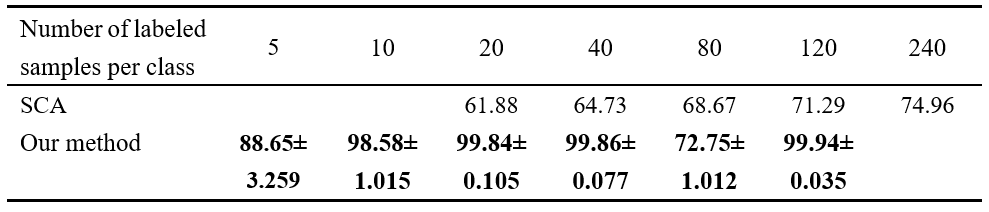}}\\
\end{tabular*}
\label{tab3}
\end{table}

2) FUSAR-ship dataset:

\begin{table}[h]
\caption{Comparison of recognition accuracy (\%) between the proposed method and other algorithms on the FUSAR-ship dataset, best results in bold.}
\label{table4}
\tablefont
\begin{tabular*}{20pc}{@{}c@{}}
\centerline{\includegraphics[width=18.5pc]{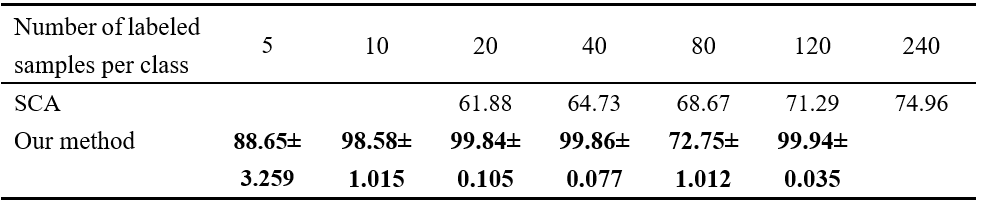}}\\
\end{tabular*}
\label{tab4}
\end{table}

\subsection{Ablation experiments}

In this subsection, ablation experiments will be conducted to provide a more intuitive demonstration of the contribution of the three components in the proposed method, including: the energy-based intra-class pseudo-label selection mechanism, the adaptive margin loss function, and the adaptive hard triplet loss function. Unless otherwise specified, all ablation experiments are conducted on the MSTAR dataset with IR set to 10 and the FUSAR-ship dataset with IR set to 10, with the proportion of labeled samples set to 20\%. The results of all ablation experiments are shown in Table 5 where the first row represents the recognition results of FixMatch ensemble with EUAPS.

1)	ESIDPS

First, the effectiveness of the energy-based intra-class pseudo-label selection mechanism is evaluated relative to the traditional method based on softmax confidence scores. In the FixMatch model, only the method based on softmax confidence scores is used to select pseudo-labels. As shown in the first row of Table 5, the FixMatch ensemble with EUAPS achieves a recognition accuracy of 93.56\% on the MSTAR dataset. When the ESIDPS mechanism is introduced, a recognition rate of 94.79\% is achieved, which is an improvement of 1.23\% compared to the baseline, as shown in the second row of Table 5. Similarly, on the FUSAR-ship dataset, the introduction of the ESIDPS mechanism increases the recognition rate from 81.74\% to 82.46\%, an improvement of 0.72\%. Thus, the energy-based intra-class pseudo-label selection mechanism effectively reduces the error rate of pseudo-labels by selecting labels close to the training distribution based on energy scores. Compared to the method based on softmax confidence scores, it exhibits higher reliability and robustness.

\begin{table}
\caption{Comparison of recognition accuracy (\%) between the proposed method and other algorithms on the FUSAR-ship dataset, best results in bold.}
\label{table5}
\tablefont
\begin{tabular*}{20pc}{@{}c@{}}
\centerline{\includegraphics[width=18.5pc]{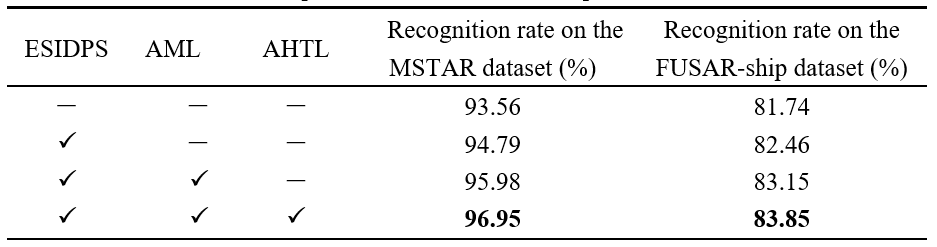}}\\
\end{tabular*}
\label{tab5}
\end{table}

2)	AML

Considering the ease of classifier-induced inter-class confusion, leading to natural imbalance in the pseudo-label generation process even when labeled and unlabeled samples are balanced in the original dataset, this study adopts an Adaptive Margin Loss (AML) to replace the cross-entropy term in traditional unsupervised losses. AML increases the margin between classification boundaries of different classes, thereby enhancing the classifier's ability to differentiate between features of different classes and counteracting inter-class confusion. As shown in the third row of Table 5 incorporating AML results in recognition accuracies of 95.98\% and 83.15\% on the MSTAR and FUSAR-ship datasets, respectively. Compared to the second row of the table, performance improves by 1.19\% and 0.69\%, respectively. Experimental results demonstrate that AML effectively alleviates bias in pseudo-labels by adaptively adjusting inter-class margins when computing unsupervised loss for pseudo-labeled samples, thereby preventing minority class samples from being incorrectly predicted as majority class, and consequently enhancing the overall model recognition rate.
To illustrate the effect of AML more intuitively, the confusion matrix obtained from pseudo-labels on the MSTAR dataset is visualized, as shown in Figure 9. Subfigures (a) and (b) in the figure represent the results obtained using cross-entropy and AML functions, respectively, to compute unsupervised loss for pseudo-labeled samples. The diagonal of the confusion matrix represents the model's correct classification probability for each class, while the numbers outside the diagonal indicate the probability of misclassification. Observing subfigure (a) of Figure 9, it can be seen that when using the cross-entropy function to compute unsupervised loss, the model incorrectly predicts some unlabeled samples of certain classes as other classes. For example, 28.8\% of samples in the "ZIL131" class are erroneously predicted as the "D7" class. However, with the introduction of the AML function, not only do the correct classification probabilities of each class improve, but also the probability of "ZIL131" class samples being misclassified as "D7" class decreases to 11.9\%, as shown in subfigure (b) of Figure 8. The analysis of the pseudo-label confusion matrix above illustrates that AML reduces inter-class confusion of the classifier, thereby counteracting the imbalance issue of pseudo-labels and effectively improving the probability of correct pseudo-label prediction.

\begin{figure}[h]
	
	\begin{minipage}{0.48\linewidth}
		\vspace{1pt}
		\centerline{\includegraphics[width=\textwidth]{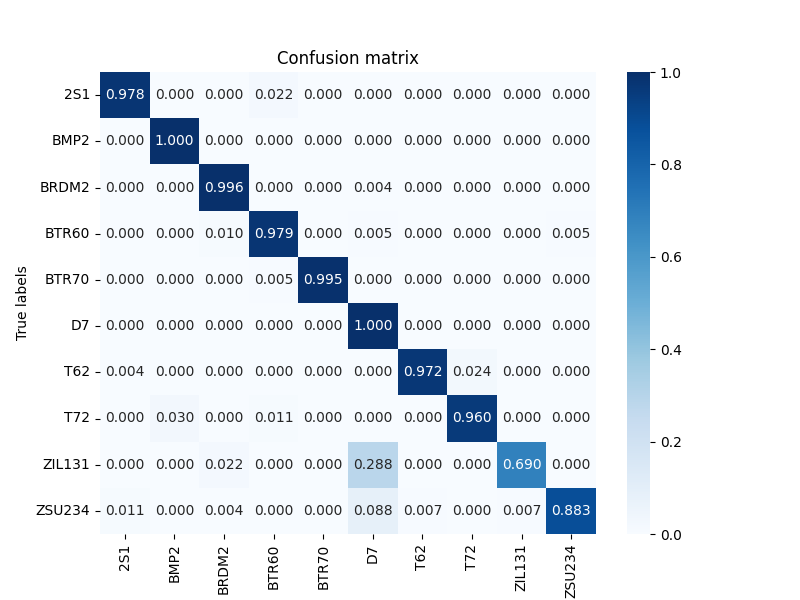}}
		\centerline{(a)	Cross-entropy function}
	\end{minipage}
	\begin{minipage}{0.48\linewidth}
		\vspace{1pt}
		\centerline{\includegraphics[width=\textwidth]{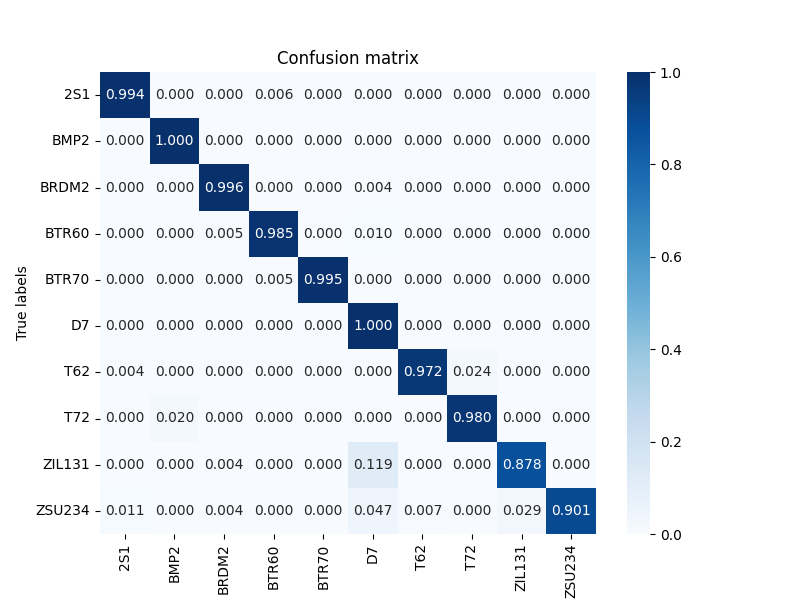}}
		\centerline{(b)	AML function}
	\end{minipage}
	\caption{The confusion matrix visualization of pseudo-labels}
	\label{fig4}
\end{figure}

3)	AHTL

The Adaptive Hard Triplet Loss (AHTL) function designed in this paper aims to construct hard triplets, directing the model's attention to more challenging samples, thereby enhancing the model's feature representation capability. Additionally, by adaptively assigning different weights to samples, it improves the robustness and stability of the model. As shown in the fourth row of Table 5, with the inclusion of AHTL, the model's recognition accuracy is further improved. Specifically, it increases by 0.97\% on the MSTAR dataset and by 0.70\% on the FUSAR-ship dataset. The improvement in recognition performance validates the effectiveness of AHTL, which positively contributes to improving recognition performance by dynamically adjusting the model's focus on hard samples during training.
Furthermore, to demonstrate that the optimized AHTL proposed in this paper is more effective in improving model recognition rate compared to traditional triplet loss and hard triplet loss, experiments were conducted on the same MSTAR and FUSAR-ship datasets, comparing them with the traditional triplet loss with uniform weights and hard triplet loss [98]. The experimental results are shown in Table 6. Observing Table 6, the proposed AHTL achieves higher recognition accuracy on both datasets compared to the two contrasted losses. On the MSTAR dataset, AHTL outperforms traditional and hard triplet losses by 2.14\% and 1.13\%, respectively. On the FUSAR-ship dataset, it surpasses them by 1.78\% and 0.91\%, respectively. It can be seen that the contribution of the proposed AHTL to the recognition model is significantly superior to the other two losses. This also confirms that AHTL is more suitable for the research in this paper and has a significant effect on addressing the bias problem of recognition models in long-tailed distribution data. Additionally, the improvement in recognition performance is also attributed to the additional distance constraints imposed by AHTL on samples with different perturbations, which helps improve the model's generalization.

\begin{table}
\caption{Comparison of recognition accuracy (\%) between the proposed method and other algorithms on the FUSAR-ship dataset, best results in bold.}
\label{table6}
\tablefont
\begin{tabular*}{20pc}{@{}c@{}}
\centerline{\includegraphics[width=18.5pc]{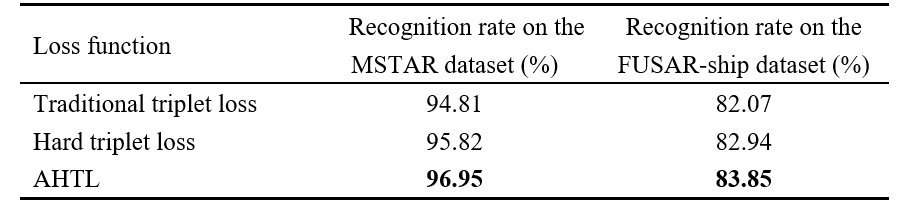}}\\
\end{tabular*}
\label{tab6}
\end{table}

\subsection{Hyper-parameter Analysis Experiment}

This section will focus on several key hyperparameters in the proposed method: the energy score threshold $\tau_{e}$ in Equation (4.7), the temperature parameter T in Equation (4.4), the distance margin m in Equation (4.11), and the loss weights $\lambda_{u}$ and $\lambda_{AHTL}$ in Equation (4.13). Unless otherwise specified, all experiments are conducted on the MSTAR dataset with IR set to 10 and the FUSAR-ship dataset, with the proportion of labeled samples set to 20%.

1)	Energy score threshold $\tau_{e}$

In the energy-based pseudo-label selection method, a key hyperparameter is the energy score threshold $\tau_{e}$ . As shown in Equation (4.7), it determines which samples can be chosen as pseudo-labels. From Equation (4.4), it can be observed that the computed energy scores are negative and proportional to the total number of classes. Additionally, through experimentation, it has been found that the energy scores of samples mainly distribute between [-11, -6]. Therefore, the energy score threshold should also be within this range. Further, by sampling different thresholds within this interval to examine their effects on the target recognition accuracy, experimental results are depicted in Figure 10. Considering that only samples with energy scores lower than $\tau_{e}$ can be selected as pseudo-labeled samples, the x-axis of Figure 10 varies from -6 to -11 to ensure that the selection process of pseudo-labels gradually becomes stricter.

From Figure 10(a), it is observed that as the threshold decreases gradually from -6 to -9.5, the recognition accuracy of the model on the MSTAR dataset steadily improves. This is because a smaller threshold makes the model more stringent in selecting pseudo-labels, only allowing unlabeled samples very close to the distribution to be chosen as pseudo-labels, thus increasing the accuracy of pseudo-labels. However, as the threshold continues to decrease within the range [-9.5, -11], the recognition accuracy shows a decreasing trend. This change occurs because some correctly predicted pseudo-labels are filtered out by the model's overly strict energy threshold, significantly affecting the recall rate of pseudo-labels and consequently impacting the overall recognition rate of the model. Similarly, observing the experimental results on the FUSAR-ship dataset shown in subfigure 10(b), when the threshold is set to -9, the recognition accuracy peaks. In summary, through the above experimental analysis, it can be determined that for the MSTAR and FUSAR-ship datasets, the optimal energy score thresholds are -9.5 and -9, respectively.

\begin{figure}[h]
	
	\begin{minipage}{0.48\linewidth}
		\vspace{1pt}
		\centerline{\includegraphics[width=\textwidth]{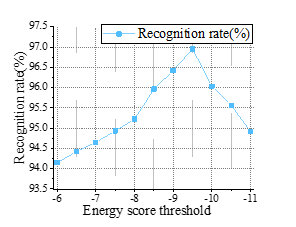}}
		\centerline{(a)	MSTAR}
	\end{minipage}
	\begin{minipage}{0.48\linewidth}
		\vspace{1pt}
		\centerline{\includegraphics[width=\textwidth]{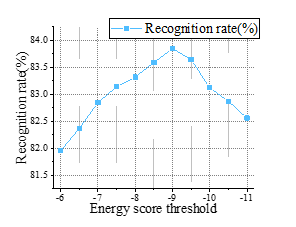}}
		\centerline{(b)	FUSAR-ship}
	\end{minipage}
	\caption{The impact of energy score threshold on recognition performance}
	\label{fig10}
\end{figure}

2)	Temperature parameter $T$

The calculation formula for the energy score includes a temperature parameter $T$, which is used to adjust the distribution of energy scores. Generally, when the temperature parameter $T$ is smaller, the differences in energy scores are larger, resulting in a steeper distribution; conversely, when $T$ is larger, the distribution becomes smoother. Here, experiments are conducted to evaluate the impact of this parameter on recognition performance. Figure 11 shows the recognition results on the MSTAR and FUSAR-ship datasets when the temperature parameter $T$ takes values {0.0, 0.5, 1.0, 1.5, 2.0, 4.0}.

\begin{figure}[h]
	
	\begin{minipage}{0.48\linewidth}
		\vspace{1pt}
		\centerline{\includegraphics[width=\textwidth]{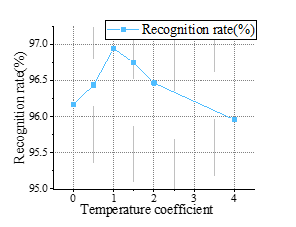}}
		\centerline{(a)	MSTAR}
	\end{minipage}
	\begin{minipage}{0.48\linewidth}
		\vspace{1pt}
		\centerline{\includegraphics[width=\textwidth]{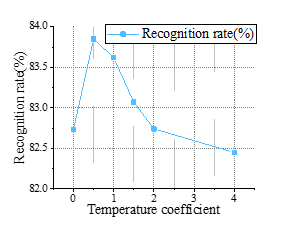}}
		\centerline{(b)	FUSAR-ship}
	\end{minipage}
	\caption{The effect of temperature coefficient on recognition performance}
	\label{fig11}
\end{figure}

Observing Figure 11(a), it can be seen that the optimal recognition performance on the MSTAR dataset is achieved when $T$ equals 1. At this point, the temperature parameter has no effect on the distribution of energy scores. However, when $T$ is greater than 1, the model's performance decreases. This is because as $T$ increases, the distribution of energy scores becomes smoother, making it more difficult to distinguish between samples from different classes. Similarly, when the temperature parameter is set to a value less than 1, the model also performs poorly. Therefore, for the MSTAR dataset, simply setting the temperature parameter $T$ to 1 is sufficient. Similarly, observing Figure 11(b), when $T$ is set to 0.5, the recognition rate peaks on the FUSAR-ship dataset. At this point, the differences in energy distribution between different ship classes increase, allowing the model to more accurately classify each class.

3)	Distance margin $m$

In the formula (4.11) of the AHTL function, there exists a crucial distance margin parameter m. Its presence constrains the distance between the anchor sample and the positive sample in a triplet, requiring it to be smaller than the distance between the anchor sample and the negative sample in the constructed feature space, thereby reducing intra-class variance and increasing inter-class variance. In this section, experiments are conducted to explore the impact of the parameter m on the recognition model and to find the optimal value.
It is clear that the distance margin parameter $m$ is a constant greater than 0. Too small values of $m$ cannot enhance the model's ability to distinguish between different class samples, while too large values increase the training difficulty of the model, leading to non-convergence. The experiments found that when the size of $m$ exceeds 0.5, the model becomes difficult to converge. Therefore, only the variation of $m$ from 0.1 to 0.4 is studied to observe the changes in the recognition results of the model on the MSTAR and FUSAR-ship datasets. The experimental results are shown in Figure 12, indicating that when the distance margin parameter $m$ is very small (e.g., 0.1), the model's recognition performance is poor on both datasets. This is consistent with the previous observation that too small values of $m$ cannot enhance the model's ability to distinguish between different class samples. As $m$ increases, the performance of the model on both datasets improves, reaching a peak when $m=0.3$. These experimental results validate that in the design of AHTL, by appropriately increasing $m$, the model's feature learning capability can be significantly improved, thus enhancing the correct recognition rate of complex samples. Overall, when $m$ is set to 0.3, the model achieves optimal performance on both the MSTAR and FUSAR-ship datasets.

\begin{figure}[h]
	
	\begin{minipage}{0.48\linewidth}
		\vspace{1pt}
		\centerline{\includegraphics[width=\textwidth]{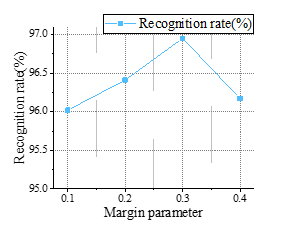}}
		\centerline{(a)	MSTAR}
	\end{minipage}
	\begin{minipage}{0.48\linewidth}
		\vspace{1pt}
		\centerline{\includegraphics[width=\textwidth]{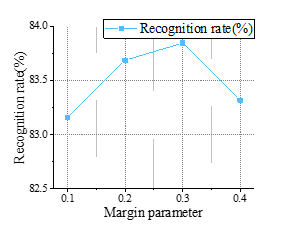}}
		\centerline{(b)	FUSAR-ship}
	\end{minipage}
	\caption{The effect of distance margin parameter on recognition performance}
	\label{fig12}
\end{figure}

4)	Loss weights $\lambda_{u}$ and $\lambda_{AHTL}$

As shown in Equation (4.13), the weights $\lambda_{u}$ and $\lambda_{AHTL}$ of the unsupervised loss function balance the supervised and unsupervised terms in the loss function. In this section, experiments are conducted on two datasets to separately discuss the effects of $\lambda_{u}$ and $\lambda_{AHTL}$ on the model performance. The experimental results are shown in Figures 13 and 14.

\begin{figure}[h]
	
	\begin{minipage}{0.48\linewidth}
		\vspace{1pt}
		\centerline{\includegraphics[width=\textwidth]{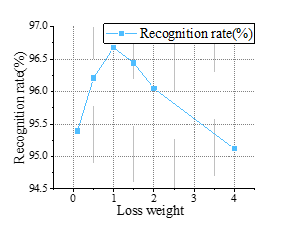}}
		\centerline{(a)	MSTAR}
	\end{minipage}
	\begin{minipage}{0.48\linewidth}
		\vspace{1pt}
		\centerline{\includegraphics[width=\textwidth]{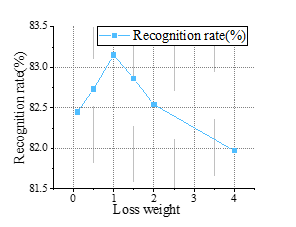}}
		\centerline{(b)	FUSAR-ship}
	\end{minipage}
	\caption{The effect of weight parameter of the unsupervised loss term on recognition performance}
	\label{fig13}
\end{figure}

\begin{figure}[h]
	
	\begin{minipage}{0.48\linewidth}
		\vspace{1pt}
		\centerline{\includegraphics[width=\textwidth]{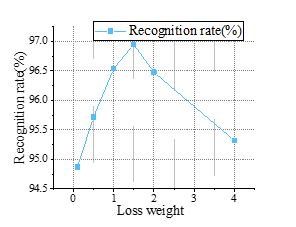}}
		\centerline{(a)	MSTAR}
	\end{minipage}
	\begin{minipage}{0.48\linewidth}
		\vspace{1pt}
		\centerline{\includegraphics[width=\textwidth]{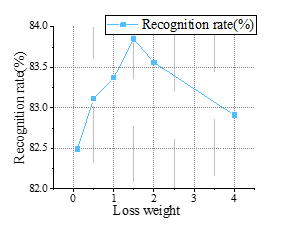}}
		\centerline{(b)	FUSAR-ship}
	\end{minipage}
	\caption{The effect of weight parameter of AHTL loss term on recognition performance}
	\label{fig14}
\end{figure}

Firstly, by fixing $\lambda_{AHTL}$ to 1, the effect of varying $\lambda_{u}$ is discussed independently. Figure 13 presents the recognition accuracy of the model on both datasets for different values of $\lambda_{u}$ within the set {0.1, 0.5, 1, 1.5, 2, 4}. It can be observed that as   gradually increases within the range [0.1, 1], the model's recognition rate also improves gradually. When $\lambda_{u}$ is set to 1, the model achieves peak performance on both the MSTAR and FUSAR-ship datasets. However, with further increases in $\lambda_{u}$, the model's performance gradually declines. Therefore, it can be determined that the optimal value of $\lambda_{u}$ is 1.

Next, with $\lambda_{u}$ fixed at 1, the optimal value of $\lambda_{AHTL}$ is investigated. The set of values for $\lambda_{AHTL}$ is {0.1, 0.5, 1, 1.5, 2, 4}. Observing Figure 14, it can be seen that when   $\lambda_{AHTL}$ is set to 0.1, the recognition performance of the model on both the MSTAR and FUSAR-ship datasets is poor. However, as $\lambda_{AHTL}$ increases within the range [0.1, 1.5], the model's recognition performance steadily improves. This change in trend is expected because the proposed Adaptive Hard Triplet Loss (AHTL) aims to enhance the model's ability to distinguish between samples from different classes. AHTL needs to have a certain weight during the training process to be effective and thus improve the overall recognition rate of the model. However, higher values of $\lambda_{AHTL}$ are not necessarily better. As shown in Figure 14, when $\lambda_{AHTL}$ exceeds 1.5, the model's performance decreases. Therefore, the optimal value of $\lambda_{AHTL}$ is determined to be 1.5. With this setting, the model in this paper can achieve optimal recognition results on both the MSTAR and FUSAR-ship datasets.

\section{Discussion}

This study proposes and validates a semi-supervised SAR target recognition method adaptable to category imbalance on two extremely imbalanced SAR target datasets, with the findings demonstrating higher recognition performance.
First, the generic softmax-based pseudo-label screening method may worsen the data imbalance issue by decreasing the pseudo-label recall of unlabeled samples in a few classes under the class imbalance scenario\cite{33,34}. In this paper, we discard the softmax-based pseudo-label selection mechanism and instead screen pseudo-labels by judging whether the unlabeled samples are close to within the training distribution. Additionally, energy scores are integrated with it for quantitative screening. The data distribution of pseudo-labels is updated and expanded during the training iteration process, increasing the pseudo-labels' dependability in situations when there is data imbalance.
When calculating the unsupervised loss of pseudo-labeling, most semi-supervised models often assign the same weight to all samples\cite{11,12,13}\cite{15}\cite{17}. While, even when the original labeled and unlabeled data are balanced, the generated pseudo-labels exhibit a natural imbalance problem\cite{35}, which is mostly caused by the classifier's inter-class confusion mistake. As a result, this research replaces an adaptive marginal loss (AML) function for the usual cross-entropy term to strengthen the model's classification marginal differences between easily confounded classes, hence mitigating the model bias problem. The experimental results of Fig. 9 reveal that the probability of the model misclassifying "D7" as "ZIL131" is greatly reduced after introducing the AML loss function, demonstrating the AML method's usefulness. However, the current research for biased pseudo-labeling is still relatively small. Therefore, further research in this area is required for semi-supervised learning in the future.
In data imbalance circumstances, machine learning algorithms are prone to bias toward majority class samples, which has a significant impact on recognition accuracy. Therefore, this work proposes an adaptable hard triple loss (AHTL) function for semi-supervised learning. The loss function not only leads the model to pay more attention to complex difficult samples during the training process, improving the network's feature learning ability, but also improves the model's stability and robustness by adaptively adjusting the weights of the difficult samples to avoid interference from noise samples and outliers. However, this study only explores how to lower model preference for most classes by modifying the model's empirical risk at the algorithmic level. Data-level approaches based on generative models have garnered a lot of attention in recent years, and they can be integrated with algorithm-level approaches. However, it is vital to emphasize that generative models are similarly sensitive to data imbalances.
In conclusion, the approach proposed in this study performs well on the unbalanced MSTAR and FUSAR-ship datasets, successfully mitigates the effects of category imbalance on the target recognition domain, and increases model recognition accuracy. Importantly, the research in this paper fills the research gap in the field of semi-supervised SAR ATR under class imbalance.

\section{Conclusions}

This work presents a semi-supervised SAR target recognition approach using dynamic energy scores and adaptive loss functions. First, the method overcomes the problem of class imbalance generated by the softmax threshold-based pseudo-label generation method, which lowers the recall rate of pseudo-labels for minority classes. Instead, a pseudo-label selection technique based on energy scores is used to generate pseudo-labels. This strategy not only improves the reliability of pseudo-labels in long-tailed distribution scenarios, but it also responds better to the constantly changing data distribution because its pseudo-labels are dynamically updated. Second, an adaptive margin-aware loss function is proposed to replace the cross-entropy term in the unsupervised loss, addressing the biased pseudo-label problem caused by the classifier's inter-class confusion. Furthermore, an adaptive hard triplet loss function is used to push the model to focus on complicated challenging samples during training, hence enhancing the model's capacity to acquire discriminative features and reducing the model's tendency for majority classes. On long-tailed distribution datasets like MSTAR and FUSAR-ship, the proposed approach significantly improves recognition performance, demonstrating its effectiveness and application. Adequate ablation and hyperparameter experiments confirm the proposed method's robustness and generalizability.

\bibliography{Energy_Score-based_Pseudo-Label_Filtering_and_Adaptive_Loss_for_Imbalanced_Semi-supervised_SAR_target_recognition}

\end{document}